\documentstyle[11pt,latex-acl]{article}

\makeatletter
\def\@eqnnum{\hbox to .01pt{}\rlap{\rm \hskip -\displaywidth(\theequation)}}

\def\examples{
 \list{*}{
  \setlength{\labelwidth}{\leftmargin}
  \setlength{\labelsep}{0 pt}
  \setlength{\rightmargin}{0 pt}
  \def\@listctr{equation}		% can't use \usecounter
  \@nmbrlisttrue			% as that initialises to zero
  \let\makelabel\@mkexlabel}}

\def\@mkexlabel#1{(\if\@itemlabel#1\@exnum%
\else #1\fi)\hfil}

\def\@exnum{{\rm \theequation}}

\newcounter{subexample}			% counter for subexamples

\def\thesubexample{\alph{subexample}}	% make them letters not numbers

\def\subexamples{
 \list{*}{
  \usecounter{subexample}
  \setlength{\rightmargin}{0 pt}
  \let\makelabel\@mksubexlabel}}

\def\@mksubexlabel#1{\if\@itemlabel#1\@subexnum%
\else #1\fi.\hfil}

\def\@subexnum{{\rm \thesubexample}}

\makeatother

\newcommand {\lsembrac}		{\mbox{[ \kern -0.5 em [}}
\newcommand {\rsembrac}		{\mbox{] \kern -0.5 em ]}}

\newcommand {\csmodels} 	{\hspace*{0.5em}\mbox{$\mid$ \kern -0.5 em
$\approx$}\hspace*{0.5em}}
\newcommand {\csmodel} 	{\mbox{$\mid$ \kern -0.5 em $\approx$}}

\newcommand {\notcsmodels} {\hspace*{0.5em}\mbox{$\mid$ \kern -0.5 em $\approx$
\kern -1 em $/$}\hspace*{0.5em}}

\newcommand {\emb} {\perp\raisebox{0.55em}{\hspace*{-1.3em}
\mbox{\rule{0.3em}{0.01in}}}\hspace*{0.15em}}

\newcommand{\wants}{{\cal W}}
\newcommand{\believes}{{\cal B}}
\newcommand{\intends}{{\cal I}}

\newcommand{\hidden}[1]{}

\title{\LARGE\bf INTENTIONS AND INFORMATION IN DISCOURSE}
\author{Nicholas Asher\\
IRIT, Universit\'{e} Paul Sabatier,\\
118 Route de Narbonne,\\
31062 Toulouse, {\sc cedex},\\
France \\
{\tt asher@irit.fr}
\And
Alex Lascarides\\
Department of Linguistics,\\
Stanford University,\\
Stanford,\\
Ca 94305-2150,\\
{\sc usa},\\
{\tt alex@csli.stanford.edu}
}

\begin{document}

\maketitle

\begin{abstract}

This paper is about the flow of inference between communicative
intentions, discourse structure and the domain during discourse
processing.
We
augment a theory of discourse interpretation with a theory of distinct
mental attitudes and reasoning about them, in order to
provide an account of how
the attitudes interact with reasoning about discourse
structure.

\end{abstract}

\section{INTRODUCTION}

The
flow of inference between
communicative intentions and domain information is often
essential to discourse
processing.  It is well reflected in this discourse from
Moore and Pollack (1992):
\begin{examples}
\item	\label{bush}
\begin{subexamples}
\item
George Bush supports big business.
\item	He's sure to veto House Bill
1711.
\end{subexamples}
\end{examples}
There are at least three different interpretations.
Consider Context 1:  in this context
the interpreter $I$ believes
that the author $A$ wants to convince him that
(\ref{bush}b) is true.
For example, the context is one in which
$I$ has already uttered {\em Bush won't veto any more bills}.
$I$ reasons that $A$'s linguistic
behavior was intentional, and therefore that $A$
believes that by saying (\ref{bush}a) he will
convince $I$ that Bush will veto the bill.  Even if $I$ believed
nothing about the bill, he now infers it's
bad for big business.
So we have
witnessed an inference from premises that involve the
desires and beliefs of $A$ (Moore and Pollack's
``intentional structure''),
as well as his linguistic behavior,
to a conclusion about domain information (Moore and Pollack's
``informational structure'').

Now consider Context 2: in this context $I$ knows
that $A$ wants to convince him of (\ref{bush}a).
As in Context 1, $I$ may infer
that the bill is bad for big business.  But now,
(\ref{bush}b) is used to
support (\ref{bush}a).

Finally, consider Context 3: in this context $I$ knows
that House Bill 1711 is bad for big business, but doesn't know $A$'s
communicative
desires prior to witnessing his linguistic behaviour.  From his beliefs
about the domain, he infers that supporting big business
would cause Bush to veto this bill.  So, $A$ must have uttered
(\ref{bush}a) to support (\ref{bush}b).  Hence $I$
realises that $A$ wanted him to believe
(\ref{bush}b).  So in contrast to Contexts 1 and 2, we have a flow of inference
from informational structure to intentional structure.

This story makes
two main points.
First, we agree with Moore and Pollack that we must
represent both the intentional import and the informational import of
a discourse.  As they show,
this is a problem
for current formulations
of Rhetorical Structure
Theory ({\sc rst}) (Thompson and Mann, 1987).
Second, we go further than Moore and Pollack, and argue that
reasoning about beliefs and desires exploits different rules and
axioms from those used to infer rhetorical relations.  Thus, we should
represent intentional structure and discourse
structure separately.  But we postulate rhetorical
relations that express the discourse
function of the constituents in the communicative plan of the author,
and we permit interaction between reasoning about rhetorical relations
and reasoning about beliefs and desires.

This paper provides the first steps towards a formal analysis of the
interaction between intentional structure and informational structure.
Our framework for discourse structure analysis is {\sc sdrt} (Asher
1993).  The basic representational structures of that theory may
be used to characterise cognitive states.  We will extend the
logical engine used to infer rhetorical relations---{\sc dice}
(Lascarides and Asher 1991, 1993a, 1993b, Lascarides
and Oberlander 1993)---to model inferences about intentional
structure and its interaction with informational structure.

\section{BUSH'S REQUIREMENTS}

We must represent both the intentional import and the informational
import of a discourse simultaneously.  So
we need a theory of discourse structure where discourse relations
central to intentional import and to
informational import can hold simultaneously between the same
constituents.  A logical
framework in which all those plausible relations between
constituents that are consistent with each other are inferred, such as
a nonmonotonic logic like that in
{\sc dice}
(Lascarides and Asher, 1993a), would achieve this.  So
conceivably, a similar nonmonotonic logic for {\sc
rst} might solve the problem of keeping
track of the intentional and informational structure
simultaneously.

But this would work only if the various
discourse relations about intentions and information could
simultaneously hold in a {\em consistent} knowledge base ({\sc kb}).
Moore and Pollack (1992) show via discourse (\ref{5}) that the current
commitment to the nucleus-satellite distinction in {\sc rst}
precludes this.
\begin{examples}
\item	\label{5}
\begin{subexamples}
\item	Let's go home by 5.
\item	Then we can get to the hardware store \\before it
closes.
\item	That way we can finish the bookshelves tonight.
\end{subexamples}
\end{examples}
 From an intentional perspective, (\ref{5}b) is a satellite to
(\ref{5}a) via {\em Motivation}.  From an
informational perspective, (\ref{5}a) is a satellite to (\ref{5}b)
via {\em Condition}.  These two
structures are incompatible.  So  augmenting {\sc rst} with a
nonmonotonic
logic for inferring rhetorical relations would not yield a
representation of (\ref{5}) on multiple
levels in which both intentional and informational relations
are represented.  In {\sc sdrt}, on the
other hand, not all discourse relations induce subordination,
and so there is more scope for different
discourse relations holding simultaneously in a consistent {\sc kb}.

Grosz and Sidner's (1986) model of discourse interpretation is
one where the same discourse elements are related simultaneously on
the informational and intentional levels.
But using their
framework to model
(\ref{bush}) is not straightforward.
As Grosz and Sidner (1990) point out:
``any model (or theory) of the communication situation must
distinguish among beliefs and intentions of different agents,'' but
theirs does not.  They represent
intentional structure as a stack of
propositions, and different attitudes aren't
distinguished.
The informal analysis  of (\ref{bush}) above demands such
distinctions, however.
For example, analysing (\ref{bush}) under Context 3 requires a
representation of the following statement:
since $A$ has provided a reason why
(\ref{bush}b) is true, he must {\em want} $I$ to {\em believe}
that (\ref{bush}b)
is true.  It's unclear how Grosz and Sidner would represent this.
{\sc sdrt} (Asher, 1993)
is in a good position to be integrated with a theory of
cognitive states, because it
uses the same basic structures (discourse representation structures
or {\sc drs}s) that have been
used in Discourse Representation Theory ({\sc drt})
to represent different attitudes like beliefs and
desires (Kamp 1981, Asher
1986, 1987, Kamp 1991, Asher and Singh, 1993).

\section{A BRIEF INTRODUCTION TO \\SDRT AND DICE}

In
{\sc sdrt} (Asher, 1993), an {\sc
nl} text is represented by a segmented {\sc drs}
({\sc sdrs}), which is a pair of
sets containing: the {\sc drs}s or {\sc sdrs}s
representing respectively sentences and text segments, and discourse
relations between them.
Discourse relations, modelled after those proposed by Hobbs (1985),
Polanyi (1985) and
Thompson and Mann (1987), link together the constituents of an {\sc
sdrs}.  We will mention three: {\em Narration},
{\em Result} and {\em Evidence}.

{\sc sdrs}s have a
hierarchical configuration, and
{\sc sdrt} predicts points of attachment in a
discourse structure for new information. Using
{\sc dice} we infer from the reader's
knowledge resources {\em which} discourse relation should be used to
do attachment.

Lascarides and Asher (1991) introduce default rules representing the
role of Gricean pragmatic maxims and domain knowledge in calculating
the value of the update function $\langle\tau,\alpha,\beta\rangle$,
which means
``the representation $\beta$ of the current sentence
is to be attached to $\alpha$ with a discourse relation, where
$\alpha$ is an open node in the representation $\tau$ of the text so
far''.  Defaults are represented by
a conditional---$\phi > \psi$ means `if
$\phi$, then normally $\psi$.  For example,
Narration says that by default {\em Narration} relates elements in a
text.
\begin{itemize}
\item	{\tt Narration}: $\langle\tau,\alpha,\beta\rangle >
Narration(\alpha,\beta)$
\end{itemize}
Associated axioms show how {\em Narration} affects the temporal
order of the events described: Narration and the corresponding
temporal
axioms on {\em Narration} predict that normally the textual order of
events
matches their temporal order.

The logic on which {\sc dice} rests is
Asher and Morreau's (1991) Commonsense Entailment ({\sc ce}).
Two
patterns of nonmonotonic inference are particularly relevant
here.  The
first is Defeasible Modus Ponens: if one default rule has its
antecedent verified, then the consequent is nonmonotonically inferred.
The second is the Penguin Principle: if there are conflicting default
rules that apply, and their antecedents are in logical entailment
relations, then the consequent of the rule with the most specific
antecedent is inferred.  Lascarides and Asher (1991) use {\sc
dice} to yield the
discourse structures and temporal
structures for simple discourses.  But the theory has so far ignored
how $A$'s intentional structure---or more accurately, $I$'s model
of $A$'s intentional structure---influences $I$'s inferences about the
domain and the discourse structure.

\section{ADDING INTENTIONS}

To discuss intentional structure, we develop a language which can
express beliefs, intentions and desires.  Following Bratman
(forthcoming) and Asher and Singh (1993), we
think of the objects of attitudes
either as plans or as
propositions.  For example,
the colloquial intention to do something---like wash
the dishes---will be expressed as an intention toward a plan, whereas
the intention that Sue be happy is an intention toward a proposition.
Plans will just consist of sequences of
basic actions $a_1; a_2;\ldots;a_n$.
Two operators---${\cal R}$ for {\em about to do} or {\em doing}, and
${\cal D}$ for
{\em having done}---will convert actions into propositions.
The attitudes we assume in our model are {\em believes}
($\believes_A\phi$ means `$A$ believes $\phi$'), {\em wants}
($\wants_A\phi$ means `$A$ wants $\phi$'), and {\em intends}
($\intends_A\phi$ means `$A$ intends $\phi$').
All of this takes place in
a modal, dynamic logic, where the propositional attitudes are supplied
with a modal semantics.  To this we add the modal conditional operator $>$,
upon which the logic of {\sc dice} is based.

Let's take a closer look at (\ref{bush}) in
Context 1.  Let the logical forms of the sentences (\ref{bush}a) and
(\ref{bush}b) be respectively $\alpha$ and $\beta$.
In Context 1, $I$
believes that $A$ wants to convince him of $\beta$ and thinks that he
doesn't believe $\beta$ already.
Following the {\sc drt} analysis of attitudes,
we assume $I$'s cognitive state has embedded in it a model
of $A$'s cognitive state, which in turn has a representation of $I$'s
cognitive state.    So $\wants_A\believes_I\beta$ and
$\believes_A\neg\believes_I\beta$ hold in $I$'s {\sc kb}.  Furthermore,
$\langle\tau,\alpha,\beta\rangle \wedge \mbox{{\em
Info}}(\alpha,\beta)$ holds in $I$'s
{\sc kb}, where $\mbox{{\em Info}}(\alpha,\beta)$
is a gloss for the semantic content of $\alpha$ and $\beta$ that $I$
knows about.\footnote{This doesn't necessarily include that
House Bill 1711 is bad for big business.}  $I$ must now
reason about what $A$ intended by his particular
discourse action.
$I$ is thus presented with a classical reasoning problem
about attitudes: how to derive what a person believes,
from a knowledge of what he wants and an observation of his
behaviour.
The classic means of constructing such a derivation uses  the
practical syllogism, a form of reasoning about action familiar
since Aristotle.
It expresses the following
maxim: Act so as to realize your goals {\em ceteris paribus}.

The practical syllogism is a rule of defeasible
reasoning, expressible in {\sc  ce} by means of the nonmonotonic
consequence relation  $\csmodel$.   The consequence relation
$\phi\csmodel\psi$ can be stated
directly in the object
language of {\sc ce} by a formula which we abbreviate
as $\emb(\phi,\psi)$ (Asher 1993).
We use $\emb(\phi,\psi)$
to state the practical syllogism.  First,
we define the notion that the {\sc kb} and $\phi$,
but not the {\sc kb} alone, nonmonotonically
yield $\psi$:
\begin{itemize}
\item	{\tt Definition:}\\
$\emb_{kb}(\phi,\psi) \leftrightarrow \emb(KB\wedge\phi,\psi)
\wedge \neg\emb(KB,\psi)$
\end{itemize}
The Practical Syllogism says that
if (a) $A$ wants $\phi$ but believes it's not true, and (b) he knows
that if $\psi$ were added
to his {\sc kb} it would by default make $\phi$
true eventually, then by default $A$ intends $\psi$.
\begin{itemize}
\item	{\tt The Practical Syllogism:}\\
\begin{tabular}[t]{l l}
(a) &
$(\wants_A(\phi) \wedge \believes_A(\neg\phi) \wedge$\\
(b) & $\believes_A(\emb_{kb}(\psi,\mbox{{\em
eventually}}(\phi)))) >$ \\
(c) & \hspace*{0.3in} $\intends_A(\psi)$
\end{tabular}
\end{itemize}

The Practical Syllogism enables $I$ to reason
about $A$'s cognitive state.  In Context 1, when substituting
in the Practical Syllogism $\believes_I\beta$ for $\phi$,
and $\langle\tau,\alpha,\beta\rangle \wedge \mbox{{\em
Info}}(\alpha,\beta)$ for $\psi$, we find that
clause (a) of the antecedent
to the Practical Syllogism is verified.  The conclusion
(c) is also verified, because $I$ assumes that
$A$'s discourse act was intentional.  This assumption could be expressed
explicitly as a $>$-rule, but we will not do so here.

Now, abduction (i.e., explanatory reasoning) as well as nonmonotonic deduction
is permitted on the Practical Syllogism.
So from knowing (a) and (c),
$I$ can conclude the premise (b).
We can state in {\sc ce}
an `abductive'  rule based on the Practical
Syllogism:
\begin{itemize}
\item	{\tt The Abductive Practical Syllogism 1} ({\sc aps1})\\
$(\wants_A(\phi) \wedge
\believes_A(\neg\phi) \wedge \intends_A(\psi)) >\\
\hspace*{0.5in}
B_A(\emb_{kb}(\psi,\mbox{{\em eventually}}(\phi)))$
\end{itemize}
{\sc aps1} allows us to conclude (b) when (a) and (c) of the Practical
Syllogism hold.  So,
the intended action $\psi$ must be one that $A$ believes will
eventually make $\phi$ true.

When we make the same substitutions for $\phi$ and $\psi$ in {\sc
aps1} as before,
$I$ will infer the conclusion of {\sc aps1} via Defeasible Modus Ponens:
$B_A(\emb_{kb}(\langle\tau,\alpha,\beta\rangle\wedge \mbox{{\em
Info}}(\alpha,\beta), \mbox{{\em eventually}}(B_I\beta)))$.  That is,
$I$ infers that $A$ believes that, by uttering what he did, $I$
will come to believe $\beta$.

In general, there may be a variety of alternatives that we could
use to substitute for
$\phi$ and $\psi$ in {\sc aps1}, in a given situation.  For usually,
there are choices on what can be abduced.  The problem of choice is
one that Hobbs {\em et al.} (1990) address by a complex weighting
mechanism.  We could adopt this approach here.

The Practical Syllogism and
{\sc aps1} differ in two important ways from the {\sc dice}
axioms concerning
discourse relations.
First, {\sc aps1} is motivated
by an {\em abductive} line of reasoning on a pattern of defeasible reasoning
involving cognitive states.  The {\sc dice} axioms are not.
Secondly, both the Practical Syllogism and
{\sc aps1} don't include
the discourse update function $\langle\tau,\alpha,\beta\rangle$
together with some information about the semantic content of $\alpha$
and $\beta$ in the antecedent, while this is a standard feature of the
{\sc dice} axioms for inferring discourse structure.

These two differences distinguish reasoning about
intentional structures and discourse structures.
But discourse structure is linked to intentional
structure in the following way.  The above reasoning
with $A$'s cognitive state has led $I$ to conclusions about the {\em
discourse function} of $\alpha$.  Intuitively,
$\alpha$ was uttered to support
$\beta$, or $\alpha$ `intentionally supports' $\beta$.
This idea of {\em
intentional support} is defined in {\sc dice} as follows:
\begin{itemize}
\item	{\tt Intends to Support:}\\
$\mbox{{\em Isupport}}(\alpha,\beta) \leftrightarrow
(\wants_A(\believes_I\beta) \wedge
\believes_A(\neg\believes_I\beta) \wedge \\
\hspace*{0.3in}
\believes_A(\emb_{kbh}(\langle\tau,\alpha,\beta\rangle \wedge
\mbox{{\em Info}}(\alpha,\beta),\mbox{{\em eventually}}(\believes_I\beta))))$
\end{itemize}
In words, $\alpha$ intentionally supports $\beta$ if and only if $A$
wants $I$ to believe $\beta$ and doesn't think he does so already, and
he also believes that by uttering $\alpha$ and $\beta$ together, so
that $I$ is forced to reason about how they should be attached with a
rhetorical relation, $I$ will come to believe $\beta$.

$\mbox{{\em Isupport}}(\alpha,\beta)$
defines a relationship between $\alpha$ and $\beta$
at the discourse structural level, in terms of $I$'s and $A$'s
cognitive states.  With it we
infer further information about the particular discourse relation
that $I$ should use to attach $\beta$ to $\alpha$.  $\mbox{{\em
Isupport}}(\alpha,\beta)$ provides the link between
reasoning about cognitive states and reasoning about discourse
structure.

Let us now return to the interpretation of (\ref{bush}) under Context 1.  $I$
concludes $\mbox{{\em Isupport}}(\alpha,\beta)$, because the right
hand side of the $\leftrightarrow$-condition in Intends to Support
is satisfied.  So $I$ passes
from a problem of reasoning about $A$'s intentional structure
to one of reasoning about discourse structure.
Now, $I$ should check to see whether $\alpha$ actually does lead him to
believe $\beta$.  This is a check on the coherence of discourse;
in order for an {\sc sdrs} $\tau$
to be coherent,
the discourse relations predicated of the constituents
must be satisfiable.\footnote{Asher (1993) discusses this point in
relation to {\em Contrast}: the discourse marker {\em but} is used
coherently only if the semantic content of the constituents it connects do
indeed form a contrast: compare {\em Mary's hair is black but her eyes
are blue}, with {\em ?Mary's hair is black but John's hair is
black}.}
Here, this amounts to justifying $A$'s belief
that  given the discourse context and $I$'s background beliefs of which
$A$ is aware, $I$ will arrive at the desired conclusion---that he believes
$\beta$.  So, $I$ must be able to infer a particular
discourse relation $R$
between $\alpha$ and $\beta$ that has what we will call the Belief Property:
$(\believes_I\alpha \wedge R(\alpha,\beta)) > \believes_I\beta$.
That is,  $R$ must be a relation that would indeed license $I$'s
concluding $\beta$ from $\alpha$.

We concentrate here for illustrative purposes on two
discourse relations with the
Belief Property:
$\mbox{{\em Result}}(\alpha,\beta)$ and $\mbox{{\em
Evidence}}(\alpha,\beta)$; or in other words, $\alpha$ results in
$\beta$, or $\alpha$ is evidence for $\beta$.
\begin{itemize}
\item	{\tt Relations with the Belief Property:}\\
$(\believes_I\alpha \wedge \mbox{{\em Evidence}}(\alpha,\beta) )
 > \believes_I\beta$\\
$(\believes_I\alpha \wedge\mbox{{\em Result}}(\alpha,\beta))
> \believes_I\beta$
\end{itemize}

The following axiom of Cooperation captures
the above reasoning on $I$'s part:
if $\alpha$  {\em Isupports} $\beta$, then it
must be possible to infer from the semantic content, that either
$\mbox{{\em Result}}(\alpha,\beta)$  or $\mbox{{\em
Evidence}}(\alpha,\beta)$ hold:
\begin{itemize}
\item	{\tt Cooperation:}\\
$(\mbox{{\em Isupport}}(\alpha,\beta) \wedge
\langle\tau,\alpha,\beta\rangle) \rightarrow \\
\hspace*{0.2in}
(\emb_{kb}(\langle\tau,\alpha,\beta\rangle \wedge \mbox{{\em
Info}}(\alpha,\beta),
\mbox{{\em Result}}(\alpha,\beta))  \vee \\
\hspace*{0.2in}
\emb_{kb}(\langle\tau,\alpha,\beta\rangle \wedge \mbox{{\em
Info}}(\alpha,\beta),
\mbox{{\em Evidence}}(\alpha,\beta)))$
\end{itemize}
The intentional
structure of $A$ that $I$ has inferred has restricted the candidate
set of discourse relations that $I$ can use to attach $\beta$ to
$\alpha$: he must use {\em Result} or {\em Evidence}, or both.
If $I$ can't accommodate
$A$'s intentions by doing this, then
the discourse will be incoherent.
We'll shortly show how Cooperation contributes to the explanation of
why (\ref{bush2}) is incoherent.
\begin{examples}
\item	\label{bush2}
\begin{subexamples}
\item	George Bush is a weak-willed president.
\item	?He's sure to veto House Bill 1711.
\end{subexamples}
\end{examples}

\section{FROM INTENTIONS TO INFORMATION: \\CONTEXTS 1 AND 2}

The
axioms above allow $I$ to use his knowledge of $A$'s cognitive
state, and the behaviour of $A$ that he observes, to (a) infer
information about $A$'s communicative intentions, and (b)
consequently to restrict the set of candidate discourse relations that
are permitted between the constituents.
According to Cooperation, $I$ must infer
that one of the permitted discourse relations does indeed hold.  When
clue words
are lacking, the semantic content of the constituents
must be exploited.  In certain cases, it's also necessary to infer
further information that wasn't explicitly mentioned in the discourse,
in order
to sanction the discourse relation.
For example, in (\ref{bush})
in Contexts 1 and 2, $I$ infers the bill is bad for big business.

Consider again discourse (\ref{bush}) in Context 1.
Intuitively, the reason
we can infer $\mbox{{\em Result}}(\alpha,\beta)$ in the analysis of
(\ref{bush}) is because (i) $\alpha$ entails a generic (Bush
vetoes bills that are bad for big business), and (ii)
this generic makes $\beta$
true, as long as we assume that House Bill 1711 is bad for big business.

To define the Result Rule below that captures this reasoning for
discourse attachment, we first define this generic-instance
relationship:
$\mbox{{\em instance}}(\phi,\psi)$ holds just
in case $\phi$ is
$(\forall x)(A(x) > B(x))$ and $\psi$ is $A[x/d] \wedge B[x/d]$.
For example, $\mbox{{\em bird}}(\mbox{{\em tweety}})\wedge
\mbox{{\em fly}}(\mbox{{\em tweety}})$ (Tweety is a bird and Tweety
flies) is an instance of $\forall x (\mbox{{\em bird}}(x) >
\mbox{{\em fly}}(x))$ (Birds fly).

The
Result Rule says that
if (a) $\beta$ is to be attached to $\alpha$, and
$\alpha$ was intended to support $\beta$, and (b) $\alpha$ entails a
generic, of which $\beta$ and $\delta$ form an instance, and (c)
$\delta$ is consistent with what $A$ and $I$ believe,\footnote{Or,
more accurately, $\delta$ must be consistent with what $I$ himself
believes, and what he believes that $A$ believes.  In other words,
$\mbox{{\sc kb}}_A$ is $I$'s model of $A$'s {\sc kb}.}
then normally, $\delta$ and
$\mbox{{\em Result}}(\alpha,\beta)$ are inferred.
\begin{itemize}
\item	{\tt The Result Rule:}\\
\begin{tabular}[t]{l l}
(a) & $(\langle\tau,\alpha,\beta\rangle \wedge \mbox{{\em
Isupport}}(\alpha,\beta) \wedge$ \\
(b) & $\emb_{kb\wedge\tau}(\alpha,\phi) \wedge
\emb_{kb\wedge\tau\wedge\delta}(\beta,\psi)\wedge \mbox{{\em
instance}}(\phi,\psi) \wedge$\\
(c) & $\mbox{{\em consistent}}(\mbox{{\sc kb}}_I \cup \mbox{{\sc
kb}}_A
\cup\delta))$\\
 & \hspace*{0.3in}$> (\mbox{{\em Result}}(\alpha,\beta) \wedge
\delta)$
\end{tabular}
\end{itemize}
The Result Rule does two things.  First, it allows us to infer
one discourse relation ({\em Result}) from those permitted by
Cooperation.  Second, it
allows us to infer a new piece of information $\delta$, in virtue of
which $\mbox{{\em Result}}(\alpha,\beta)$ is true.

We might want further constraints on $\delta$ than that in (c); we
might add that $\delta$ shouldn't violate expectations generated by
the text.  But note that the Result Rule doesn't choose
between different $\delta$s that verify clauses (b) and (c).  As we've
mentioned, the theory needs to be extended to deal with the problem of
choice, and it may be necessary to adopt
strategies for choosing among alternatives, which take
factors other than logical structure into account.

We have a similar rule for inferring $\mbox{{\em
Evidence}}(\beta,\alpha)$
(``$\beta$ is evidence for $\alpha$'').  The Evidence rule resembles the
Result Rule, except that the textual order of the discourse
constituents, and the direction of intentional support changes:
\begin{itemize}
\item	{\tt The Evidence Rule:}\\
\begin{tabular}[t]{l l}
(a) & $(\langle\tau,\alpha,\beta\rangle \wedge \mbox{{\em
Isupport}}(\beta,\alpha) \wedge$ \\
(b) & $\emb_{kb\wedge\tau}(\alpha,\phi) \wedge
\emb_{kb\wedge\tau\wedge\delta}(\beta,\psi)\wedge \mbox{{\em
instance}}(\phi,\psi) \wedge$\\
(c) & $\mbox{{\em consistent}}(\mbox{{\sc kb}}_I \cup \mbox{{\sc
kb}}_A \cup \delta))$\\
 & \hspace*{0.3in}$> (\mbox{{\em Evidence}}(\beta,\alpha) \wedge
\delta)$
\end{tabular}
\end{itemize}
We have seen that clause (a) of the Result Rule
is satisfied
in the analysis of (\ref{bush}) in Context 1.
Now, let $\delta$ be the proposition that the House Bill 1711 is bad for
big business (written as $\mbox{{\em bad}}(1711)$).
This is consistent with
$\mbox{{\sc kb}}_I\cup \mbox{{\sc kb}}_A$, and so clause (c) is satisfied.
Clause (b) is also satisfied, because (i) $\alpha$
entails Bush vetoes bills that are bad for big business---i.e.,
$\emb_{KB\wedge \tau}(\alpha,\phi)$ holds, where $\phi$ is $\forall
x((\mbox{{\em
bill}}(x)
\wedge bad(x)) > veto(bush,x))$; (ii)
$\beta \wedge \delta$ is $bill(1711) \wedge veto(bush,1711)
\wedge bad(1711)$; and so (iii)
$\mbox{{\em instance}}(\phi,\beta\wedge\delta)$ and
$\emb_{KB\wedge\tau\wedge\delta}(\beta,\beta\wedge\delta)$ both hold.

So, when interpreting (\ref{bush}) in Context 1, two rules apply:
Narration and the Result Rule.
But the consequent of Narration
already conflicts with what is known; that the discourse relation
between $\alpha$ and $\beta$ must satisfy the Belief
Property.
So
the consequent of the Result Rule is
inferred: $\delta$ (i.e.,
House Bill 1711 is bad for big business) and $\mbox{{\em
Result}}(\alpha,\beta)$ .\footnote{
We could have a similar rule to the Result Rule for inferring
$\mbox{{\em Evidence}}(\alpha,\beta)$ in this discourse context too.}

These rules show how (\ref{bush}) can make the knowledge that the
house bill is bad for big business moot; one does not need to know
that the house bill is bad for big business
prior to attempting discourse attachment.
One can infer it at the time when discourse
attachment is attempted.

Now suppose that we start from different premises, as provided by Context
2: $\believes_A\believes_I\beta$,
$\believes_A\neg\believes_I\alpha$ and $\wants_A\believes_I\alpha$.
That is, $I$ thinks $A$ believes that $I$ believes Bush will veto the
bill, and $I$ also thinks that $A$ wants to convince him that Bush
supports big business.
Then the `intentional' line of reasoning yields different results from
the same observed behaviour---$A$'s utterance of (\ref{bush}).
Using {\sc aps1} again, but substituting $\believes_I\alpha$ for
$\phi$ instead of $\believes_I\beta$, $I$ concludes
$\believes_A(\emb_{kb}(\langle\tau,\alpha,\beta\rangle \wedge \mbox{{\em
Info}}(\alpha,\beta),\mbox{{\em
eventually}}(\believes_I\alpha))$.\footnote{Given the new {\sc kb},
the antecedent of {\sc aps1} would no longer be verified if we
substituted $\phi$ with $\believes_I\beta$.}
So $\mbox{{\em Isupports}}(\beta,\alpha)$ holds.
Now the antecedent to Cooperation is verified, and so
in the monotonic component of {\sc ce}, we infer that
$\alpha$ and $\beta$ must be connected by a discourse relation $R'$
such that $(\believes_I\beta \wedge R'(\alpha,\beta)) > \believes_I\alpha$.  As
before, this restricts the set of permitted discourse relations for
attaching $\beta$ to $\alpha$.  But unlike before, the textual order
of $\alpha$ and $\beta$, and their direction of intentional support
mismatch.  The rule that applies this time is the Evidence Rule.
Consequently, a different discourse relation is inferred, although the
same information $\delta$---that House Bill 1711 is bad for
big business---supports the discourse relation, and
is also be inferred.

In contrast,
the antecedents of the Result and Evidence Rules aren't verified
in (\ref{bush2}).  Assuming $I$
knows about the legislative process, he knows
that if George Bush is a weak willed president, then
normally, he won't veto bills.   Consequently, there is no $\delta$ that
is consistent with his {\sc kb}, and sanctions the {\em Evidence} or
{\em Result} relation.  Since $I$ cannot infer
which of the permitted discourse relations holds, and so
by contraposing the axiom Cooperation, $\alpha$ doesn't
{\em Isupport} $\beta$.  And so $I$ has failed to conclude what $A$
intended by his discourse
action.  It can no longer be a belief that it will eventually lead to
$I$ believing $\beta$, because otherwise $\mbox{{\em
Isupport}}(\alpha,\beta)$ would be true via the rule
Intends To Support.  Consequently, I cannot infer
what discourse relation to use in attachment, yielding
incoherence.

\section{FROM INFORMATION TO INTENTIONS: \\CONTEXT 3}

Consider the interpretation of (\ref{bush}) in Context 3:
$I$ has no knowledge of $A$'s communicative intentions prior
to witnessing his linguistic behaviour, but he does know that the
House Bill 1711 is bad for big business.
$I$ has sufficient information about the
semantic content of $\alpha$ and $\beta$ to infer
$\mbox{{\em Result}}(\alpha,\beta)$, via a rule given in Lascarides
and Asher (1991):
\begin{itemize}
\item	{\tt Result}\\
$(\langle\tau,\alpha,\beta\rangle \wedge \mbox{{\em
cause}}(\alpha,\beta)) > \mbox{{\em Result}}(\alpha,\beta)$
\end{itemize}
$\mbox{{\em Result}}(\alpha,\beta)$ has
the Belief Property, and $I$ reasons that from believing $\alpha$, he will
now come to believe $\beta$.  Having used the information structure to
infer discourse structure, $I$ must now come to some conclusions about
$A$'s cognitive state.

Now suppose that $\believes_A\believes_I\alpha$ is in $I$'s {\sc kb}.  Then
the following principle of Charity allows $I$ to assume that $A$ was
aware that $I$ would come to believe $\beta$ too, through doing
the discourse
attachment he did:
\begin{itemize}
\item	{\tt Charity:} $\believes_I\phi > \believes_A\believes_I\phi$
\end{itemize}
This is because $I$ has inferred $\mbox{{\em Result}}(\alpha,\beta)$,
and since {\em Result} has the belief property, $I$ will come to
believe $\beta$ through believing $\alpha$; so substituting $\beta$
for $\phi$ in Charity, $\believes_A\believes_I\beta$ will become part
of $I$'s {\sc kb} via Defeasible Modus Ponens.
So, the following is now part of $I$'s {\sc kb}:\\
$\believes_A(\emb_{kb}(\langle\tau,\alpha,\beta\rangle \wedge
\mbox{{\em Info}}(\alpha,\beta)),\mbox{{\em eventually}}(\believes_I\beta))$.
Furthermore, the assumption that $A$'s discourse behaviour was
intentional again yields the following as part of $I$'s {\sc kb}:
$\intends_A(\langle\tau,\alpha,\beta\rangle \wedge \mbox{{\em
Info}}(\alpha,\beta))$.  So, substituting
$\believes_I\beta$ and $\langle\tau,\alpha,\beta\rangle \wedge
\mbox{{\em Info}}(\alpha,\beta)$ respectively for $\phi$ and
$\psi$ into the Practical
Syllogism, we find that clause (b) of the premises, and
the conclusion are verified.  Explanatory reasoning on the Practical
Syllogism this time permits us to infer clause (a):
$A$'s communicative goals were to convince $I$
of $\beta$, as required.

The inferential mechanisms going from discourse structure to
intentional structure are much less well understood.   One
needs to be able to make some suppositions about the beliefs of $A$
before one can infer anything about his desires to
communicate, and this requires a general
theory of commonsense belief attribution on the basis of beliefs
that one has.

\section{IMPERATIVES AND\\ PLAN UPDATES}

The
revision of intentional structures exploits modes of speech other
than the assertoric.  For instance, consider another discourse from
Moore and Pollack (1992):
\begin{examples}
\item	[\ref{5}]
\begin{subexamples}
\item	Let's go home by 5.
\item	Then we can get to the hardware store \\before it
closes.
\item	That way we can finish the bookshelves tonight.
\end{subexamples}
\end{examples}

Here, one exploits how the imperative
mode affects reasoning about intentions.
Sincere Ordering captures the intuition that if $A$ orders
$\alpha$, then normally he wants $\alpha$ to be true; and
Wanting and Doing captures the intuition that if $A$ wants $\alpha$ to
be true, and doesn't think that it's impossible to bring $\alpha$ about,
then by default he intends to ensure that $\alpha$ is brought about,
either by doing it himself, or getting someone else to do it (cf.
Cohen and Levesque, 1990a).
\begin{itemize}
\item	{\tt Sincere Ordering:}\\
$Imp(\alpha) > \wants_A(\alpha)$.
\item	{\tt Wanting and Doing:}\\
$(\wants_A\alpha \wedge \neg \believes_A \neg \mbox{{\em
eventually}}({\cal R}\alpha)) > \intends_A({\cal R}\alpha)$
\end{itemize}

These rules about $A$'s intentional structure help us analyse
(\ref{5}).  Let the logical forms of (\ref{5}a-c) be
respectively $\alpha$, $\beta$ and $\gamma$.
Suppose that we have inferred by the linguistic clues
that $\mbox{{\em Result}}(\alpha,\beta)$ holds.  That is, the action
$\alpha$ (i.e., going home by 5pm), results in $\beta$ (i.e.,
the ability to go to the hardware store before it closes).
Since $\alpha$ is an imperative, Defeasible
Modus Ponens on Sincere Ordering yields the inference
that $\wants_A\alpha$ is true.  Now let us assume that the interpreter $I$
believes that the author $A$ doesn't believe that $\alpha$'s being
brought about is impossible.  Then we may use Defeasible
Modus Ponens again on Wanting and Doing, to infer
$\intends_A({\cal R}\alpha)$.  Just how the interpreter comes to the
belief, that the author believes $\alpha$ is possible,
is a complex matter.  More than
likely, we would have to encode within the extension of {\sc dice} we
have made, principles that are familiar from autoepistemic reasoning.
We will postpone this exercise, however, for another time.

Now, to connect intentions and plans with discourse structure, we
propose a rule that takes an author's use of a particular discourse
structure to be {\em prima facie} evidence that the author has a
particular intention.  The rule Plan Apprehension below, states that if
$\alpha$ is a plan that $A$ intends to do, or get someone else to do,
and he states that $\delta$ is possible as a {\em Result} of this
action $\alpha$, then the interpreter may normally take the author $A$
to imply that he intends $\delta$ as well.
\begin{itemize}
\item	{\tt Plan Apprehension:}\\
$(\mbox{{\em
Result}}(\alpha,\beta) \wedge \intends_A({\cal R}\alpha) \wedge
\beta = \mbox{{\em can}}(\delta)) > \intends_A({\cal R}(\alpha;\delta))$
\end{itemize}
We call this rule Plan
Apprehension, to make clear that it furnishes one way for the
interpreter of a verbal message, to form an idea of the author's
intentions, on the basis of that message's discourse structure.

Plan Apprehension uses
discourse structure to attribute complex plans to $A$.
And when attaching $\beta$ to $\alpha$, having inferred
$\mbox{{\em Result}}(\alpha,\beta)$, this rule's antecedent is
verified, and so we infer that $\delta$---which in this case
is to go to the hardware store before it closes---as part of
$A$'s plan, which he intends to bring about, either himself, or by
getting another agent to do it.

Now, we process $\gamma$.  {\em That way} in
$\gamma$ invokes an anaphoric
reference to a complex plan.  By the accessibility
constraints in {\sc sdrt}, its antecedent
must $[\alpha;\delta]$, because this is the only plan in the
accessible
discourse context.    So $\gamma$ must be the {\sc drs} below: as a
result of doing this plan, finishing the
bookshelves (which we have labelled $\epsilon$) is possible:
\begin{examples}
\item
[$\gamma$]
$\mbox{{\em Result}}([\alpha;\delta], \mbox{{\em
can}}(\epsilon))$
\end{examples}

Now, substituting $[\alpha;\delta]$ and $\epsilon$ for $\alpha$ and
$\beta$
into the
Plan Apprehension Rule, we find that the antecedent to this
rule is verified again, and so its consequent is nonmonotonically
inferred:  $\intends_A({\cal R}(\alpha;\delta;\epsilon))$.  Again, $I$ has used
discourse structure to attribute plans to $A$.

Moore and Pollack (1992) also discuss one of $I$'s possible responses to
(\ref{5}):
\begin{examples}
\item	\label{saw}
We don't need to go to the hardware store.  \\I borrowed a saw from
Jane.
\end{examples}

Why does $I$ respond with (\ref{saw})?  $I$ has inferred the existence
of the plan
$[\alpha;\delta;\epsilon]$ via Plan Apprehension; so he
takes the overall goal of $A$ to be $\epsilon$ (to finish the
bookshelves this evening).  Intuitively, he fills in $A$'s plan with
the reason why going to the hardware store is a subgoal:
$I$ needs a saw.  So $A$'s plan is augmented with another subgoal
$\zeta$,
where $\zeta$ is to buy a saw, as follows:
$\intends_A({\cal R}[\alpha;\delta;\zeta;\epsilon])$.
But since $\zeta$ holds, he says this and assumes that this
means that $A$ does not have to do $\alpha$ and $\delta$
to achieve $\zeta$.
To think about this formally, we need to not
only reason about intentions but also how agents update their
intentions or revise them when presented with new information.  Asher
and Koons (1993) argue that the following schema captures part of the
logic which underlies updating intentions:
\begin{itemize}
\item	$\mbox{{\em Update}}\intends_A({\cal
R}[\alpha_1;\ldots;\alpha_n]),
{\cal
D}(\alpha_1;\ldots;\alpha_j) \models\\
\hspace*{0.3in}\intends_A({\cal R}[\alpha_{j+1};\ldots;\alpha_n]) \wedge \neg
\intends_A({\cal R}[\alpha_1;\ldots;\alpha_j])$\\
\end{itemize}

In other words, if you're updating your intentions to do
actions $\alpha_1$ to
$\alpha_n$, and $\alpha_1$ to $\alpha_j$ are already done, then the
new intentions are to do $\alpha_{j+1}$ to $\alpha_n$, and you no
longer intend to do $\alpha_1$ to $\alpha_j$.

The question is now: how does this interact with discourse
structure?  $I$ is attempting to be helpful to $A$; he
is trying to help realize $A$'s goal.  We
need axioms to model this.
Some key tools for doing this have been developed in the
past couple of decades---belief revision, intention and plan
revision---and the long term aim would be to enable
formal theories of discourse structure to interact with these formal
theories of attitudes and
attitude revision.  But since a clear understanding of how intentions
are revised is yet to emerge,
any speculation on the revision of intentions in a particular
discourse context seems premature.

\newpage
\section{CONCLUSIONS AND\\ FURTHER WORK}

We have argued that it is important to separate reasoning about mental
states
from reasoning about discourse structure,
and we have suggested how to integrate a formal theory
of discourse attachment with commonsense reasoning about the discourse
participants' cognitive states and actions.

We exploited a classic
principle of commonsense reasoning about action,
the Practical Syllogism, to model $I$'s inferences about
$A$'s cognitive state during discourse processing.
We also showed how axioms could be
defined, so as to enable information to mediate between
the domain, discourse structure and communicative intentions.

Reasoning about intentional structure took a different form from
reasoning about discourse attachment, in that explanatory reasoning or
abduction was permitted for the former but not the latter (but cf.
Hobbs {\em et al}, 1990).  This, we argued, was a principled reason
for maintaining separate representations of intentional structure and
discourse structure, but preserving close links between them via
axioms like Cooperation.  Cooperation enabled $I$ to use $A$'s
communicative intentions to reason about discourse relations.

This paper provides an analysis of only very simple discourses, and
we realise that although we have introduced distinctions among the
attitudes, which we have exploited during discourse processing, this
is only a small part of the story.

Though {\sc dice} has used domain specific
information to infer discourse relations, the rules relate domain
structure to discourse structure in at best an indirect way.
Implicitly, the use of the discourse update function
$\langle\tau,\alpha,\beta\rangle$ in the {\sc dice} rules
reflects the intuitively obvious fact that domain information is
filtered through the cognitive state of $A$.  To make this explicit,
the discourse community should integrate work on speech acts and
attitudes (Perrault 1990, Cohen and Levesque 1990a, 1990b)
with theories of discourse structure.
In future work, we will investigate discourses where other axioms
linking the different attitudes and discourse
structure are important.

\section{REFERENCES}

Asher, Nicholas (1986) Belief in Discourse Representation Theory, {\em
Journal of Philosophical Logic}, {\bf 15}, 127--189.

Asher, Nicholas (1987) A Typology for Attitude Verbs, {\em Linguistics
and Philosophy}, {\bf 10}, pp125--197.

Asher, Nicholas (1993) {\em Reference to Abstract Objects in
Discourse}, Kluwer Academic Publishers, Dordrecht, Holland.

Asher, Nicholas and Koons, Robert (1993) The Revision of Beliefs and
Intentions in a Changing World, in {\em Precedings of the {\sc aai}
Spring Symposium Series: Reasoning about Mental States:  Formal
Theories and Applications}.

Asher, Nicholas and Morreau, Michael (1991)  Common Sense Entailment: A Modal
Theory of Nonmonotonic Reasoning, in {\em Proceedings to the 12th
International Joint Conference on Artificial Intelligence},
Sydney Australia, August 1991.

Asher, Nicholas and Singh, Munindar (1993) A Logic of Intentions and
Beliefs, {\em Journal of Philosophical Logic}, {\bf 22} 5, pp513--544.

Bratman, Michael (forthcoming) {\em Intentions, Plans and Practical
Reason}, Harvard University Press, Cambridge, Mass.

Cohen, Phillip R. and Levesque, Hector J. (1990a) Persistence, Intention,
and Commitment,
In Philip R. Cohen, Jerry Morgan and Martha E. Pollack (editors)
{\em Intentions in Communication}, pp33--69.
Cambridge, Massachusetts: Bradford/MIT Press.

Cohen, Phillip R. and Levesque, Hector J. (1990b) Rational Interaction
and the Basis for Communication,
In Philip R. Cohen, Jerry Morgan and Martha E. Pollack (editors)
{\em Intentions in Communication}, pp221--256.
Cambridge, Massachusetts: Bradford/MIT Press.

Grosz, Barbara J. and Sidner, Candice L. (1986) Attention, Intentions and
the Structure of Discourse.  {\em Computational Linguistics}, {\bf
12}, 175--204.

Grosz, Barbara J. and Sidner, Candice L. (1990)
Plans for Discourse.
In Philip R. Cohen, Jerry Morgan and Martha E. Pollack (editors)
{\em Intentions in Communication}, pp417--444.
Cambridge, Massachusetts: Bradford/MIT Press.

Hobbs, Jerry R. (1985) On the Coherence and Structure of Discourse.
Report No: CSLI-85-37, Center for the Study of Language and
Information, October 1985.

Kamp, Hans (1981) A Theory of Truth and Semantic Representation,
in Groenendijk, J. A. G., Janssen, T. M. V., and Stokhof, M. B. J.
(eds.) {\em Formal Methods in the Study of Language}, 277--332.

Kamp, Hans (1991) Procedural and Cognitive Aspects of
Propositional Attitude Contexts, Lecture Notes from the Third European
Summer School in Language, Logic and Information, Saarbr\"{u}cken, Germany.

Lascarides, Alex and Asher, Nicholas (1991) Discourse Relations and
Defeasible Knowledge, in {\em Proceedings of the 29th Annual Meeting
of Computational Linguistics}, 55--63, Berkeley California, {\sc usa},
June 1991.

Lascarides, Alex and Asher, Nicholas (1993a) Temporal Interpretation,
Discourse Relations and Commonsense Entailment, in
{\em Linguistics and Philosophy}, {\bf 16}, pp437--493.

Lascarides, Alex and Asher, Nicholas
(1993b) A Semantics and Pragmatics for the
Pluperfect, in {\em Proceedings of the
European Chapter of the Association for
Computational Linguistics (EACL93)}, pp250--259, Utrecht, The Netherlands.

Lascarides, Alex, Asher, Nicholas and Oberlander, Jon (1992) Inferring
Discourse Relations in Context, in {\em Proceedings of the 30th Annual
Meeting of the Association of Computational Linguistics}, pp1--8, Delaware
USA, June 1992.

Lascarides, Alex and Oberlander, Jon (1993) Temporal Connectives in a
Discourse Context, in {\em Proceedings of the
European Chapter of the Association for
Computational Linguistics (EACL93)}, pp260--268, Utrecht, The
Netherlands.

Moore, Johanna and Pollack, Martha (1992) A Problem for RST: The Need for
Multi-Level Discourse Analysis {\em Computational Linguistics}, {\bf
18} 4, pp537--544.

Perrault, C. Ray (1990) An Application of Default Logic to Speech
Act Theory,
in Philip R. Cohen, Jerry Morgan and Martha E. Pollack (editors)
{\em Intentions in Communication}, pp161--185.
Cambridge, Massachusetts: Bradford/MIT Press.

Polanyi, Livia (1985) A Theory of Discourse Structure and Discourse
Coherence, in Eilfor, W. H., Kroeber, P. D., and Peterson, K. L.,
(eds), {\em Papers from the General Session a the Twenty-First
Regional Meeting of the Chicago Linguistics Society}, Chicago, April
25--27, 1985.

Thompson, Sandra and Mann, William (1987) Rhetorical Structure Theory: A
Framework for the Analysis of Texts.  In {\em IPRA Papers in
Pragmatics}, {\bf 1}, 79--105.

\end{document}